\documentclass[journal,comsoc]{IEEEtran}

\usepackage[colorlinks]{hyperref} 
\usepackage[T1]{fontenc}

\usepackage{graphicx} 							
\usepackage{hyperref} 							
\usepackage{mathrsfs}
\usepackage[noadjust]{cite}
\usepackage{amsmath}

\begin{document}

\title{Training Neural Networks Based on Imperialist Competitive Algorithm for Predicting \\ Earthquake Intensity}

\author{Mohsen Moradi
\thanks{M. Moradi was with the Department of Mathematics, Amirkabir University of Technology - Tehran Polytechnic , Tehran, Iran, ( e-mail:
m.moradi8111@aut.ac.ir ).}}

\maketitle

\begin{abstract}
In this study we determined neural network weights and biases by Imperialist Competitive Algorithm (ICA) in order to train network for predicting earthquake intensity in Richter. For this reason, we used dependent parameters like earthquake occurrence time, epicenter's latitude and longitude in degree, focal depth in kilometer, and the seismological center distance from epicenter and earthquake focal center in kilometer which has been provided by Berkeley data base. The studied neural network has two hidden layer: its first layer has 16 neurons and the second layer has 24 neurons. By using ICA algorithm, average error for testing data is 0.0007 with a variance equal to 0.318. The earthquake prediction error in Richter by MSE criteria for ICA algorithm is 0.101, but by using GA, the MSE value is 0.115. 
\end{abstract}


\IEEEpeerreviewmaketitle

\section{Introduction}

\IEEEPARstart{T}{h}E Artificial Neural Network (ANN) is constructed of some layers and nodes. Any network should have an output and input layers, and some other hidden networks which can be added. Any layer has some nodes which are known as neurons.  Relationship between layers are based on connection between neurons. For any of these connections a weight is specified. Moreover, for any neuron, an independent parameter by the name of bias is defined. Weights and biases of the next layers are effected by a transition function, which can be linear or nonlinear and is defined by user. Neural networks have many uses in image processing, communication theory, and finance. Kaastra et al. used neural networks for forecasting time series data \cite{Kaastra}. In the study by Shaohui et al., (2003), neural networks are used in steganography for detecting hidden images which this is done by finding statistical features \cite{shaohui2003neural}. Steganography is hiding secretive information in other images for deceiving people e.g. concealing substantial features of face images like texture in other pictures \cite{moradi2017combining}. One of the main issues of neural network is determining proper values for weights and biases which in this paper, we use ICA for prognosticating earthquake intensity.  

\section{	Related works }

In \cite{otari2012review}, for detecting and forecasting natural disasters like earthquake, some data mining techniques have been used. By analyzing data mining approaches like logical models, neural networks, Bayesian networks, and decision tree, it has been illustrated that all of these methods can be used for prognosticating earthquake, tsunami, landslides, and other microseisms. 

Negarestani used neural networks for measuring Radon density in soil by local parameters \cite{negarestani2002layered}. Their analysis illustrates that by neural network approach unnatural phenomenon on earth like earthquake can be distinguished by soil's Radon density. 

In another work Hanna used a General Regression Neural Network (GRNN) for estimating soil slide potential \cite{hanna2007neural}. Their data for training network belonged to two disastrous earthquakes in Turkey and Taiwan in 1999. In their research, 620 data set including twelve soil and earthquake parameters have been introduced. For enhancing accuracy, an iterative process has been applied. 

Next, Lin et al. by using a neural network approach and 955 data set related to Alishan's neighbourhood highways in Taiwan, created a model for research on earthquake \cite{lin2009neural}. 

Another research done in this area is about assessing structures vulnerability against earthquake \cite{koutsourelakis2010assessing}. In this study, a Bayesian framework has been introduced which can produce estimates for curves friability derivatives without considering the data values. Moreover, by using Markov Chain Monte Carlo (MCMC) in their methods, they have improved the estimates. 

Slope stability during an earthquake is an important subject in geotechnical engineering. Gordan et al. studied forecasting seismic slope stability by a hybrid Artificial Neural Network and Particle Swarm Optimization \cite{gordan2016prediction}. In this study, features which have an effect on safety factor are slope heights, cohesion, and friction angle of earth. 

Baddari and et.al. have used Radial Basis Function (RBF) for studying seismic data to identify and eliminate accidental noises which have been produced \cite{baddari2009application}. For avoiding local minimums, they used a backward propagation algorithm for training network. 

\section{Imperialist Competitive Algorithm}

Atashpaz et al. have introduced ICA as an optimization goal for modeling processes \cite{atashpaz2007imperialist}. The members of initial population in this algorithm are known as countries that are produced randomly. Some powerful countries which are called imperialist colonize weak countries. The population of these colonized countries will be divided to some empires. Each empire has one imperialist and some colonies. The power of any empire is the sum of the imperialist power and a portion of its colonies powers. When the empires have been stablished, they will compete against each other. The weakest empires would be decomposed, and their colonies would join to other empires as we can see in the history. In nut shell, the strong empires would be more powerful as the result of topple of other empires. Any imperialist would help to their colonies to become stronger. The final step of algorithm is when we have only one empire in all over the world and its colonies power is close to that of imperialist. 

Imperialists would attract the colonies toward themselves by helping them in some ways like improving their infrastructures and universities. In any iteration of algorithm, a colony will move toward imperialist by X unit which is a random number by uniform or arbitrary distribution. Moreover, it is possible that in any movement of colonies toward imperialist they become more powerful than their imperialist. In this case, algorithm would choose that colony as a new imperialist, and the old imperialist would become its colony. As a result, other colonies would move toward this new imperialist in the empire. 

\section{Genetic Algorithm}
Genetic Algorithm (GA) is an optimization method which had introduced by John H. Holland in the 1970s \cite{holland1992adaptation}. This algorithm is inspired by natural evolution laws in following steps:\\

a)	All the creatures struggle against each other for being alive and those who are stronger have more chance to remain alive and make an adaptation. \\

b)	Those with high ability for adaptation survive and produce next generation. This next generation will be more advance than the previous one. \\

c)	Children are similar to their parents but they are not identical with them. Since not only do they inherit their parents’ features, but also, due to genetic mutation, they might become more improved than their parents.\\

d) After some generation, children have more adaption to the nature. \\

Genetic algorithm does the same thing in a way that for implementing the algorithm, we should define the population size, how they breed, and the rate of adaptation. Defining population is based on chains composed of genes which makes GA implementing complicated. Parameters like population size, mutation rate, and crossover rate are control parameters that should be specified before running the algorithm. Moreover, for ending the process of producing the next generation, the algorithm needs an ending condition which can be the specific number of generations or minimum amount of an optimization value. The genetic algorithm steps are as follows:

	 \subsection*{$\bullet$ Selection}
In this step, among the available chromosomes, some of them are chosen to be used in producing the next generation. Better chromosomes have more chance for selection. 

 \subsection*{$\bullet$ Crossover}
In this step, a pair of parental chromosomes generates a new pair. Only a specific kind of parents are chosen for crossover with a probability of $P_{c}$. 

 \subsection*{$\bullet$ Mutation}
After the crossover operation, mutation is done on the chromosomes. In this step, a chromosome's gene is selected by accident and it will be changed. The probability of mutation action on any chromosome is mutation rate and is shown by $P_{m}$ .\\

After mutation, the produced chromosomes, known as the new generation, will be used for the next round of algorithm.

\section{Used neural network}
For $N$ points in a page like $(x_{i},y_{i})$, $1\leq i \leq N$ which are approximately around a straight line, a good approximation can be a linear line. For finding this line, the coefficients $(a^{\ast},b^{\ast})$ should be so that we obtain  $y=a^{\ast} x+b^{\ast}$. An appropriate criterion for finding suitable line is the Minimum Mean Square Error (MMSE) which means the distance of points from that line should be minimized as the equation \ref{MMSE}.
\begin{equation} \label{MMSE}
\min_{a,b} \sum_{i=1}^{N} (e_{i})^{2}
\end{equation}

On the other hand, if points be scattered around a cubic curve, we need at least three parameters like $(a^{\ast}, b^{\ast} ,c^{*})$ to approximate the curve properly. If we use a linear line, the mean square error would be far higher than what is acceptable. \\
In a situation where the points on the plane do not have a regular pattern and have a lot of maximum and minimum points, high degree curves should be used for approximation. Moreover, if the number of points are not enough or the points on plane are in high dimensional surfaces, the problem would be doubled. Therefore, we should find another solution. To overcome this, we would use Multilayer Perceptron (MLP) neural network. Generally, this neural networks are constructed by some layers which any layer also has some number of neurons. 

Any neuron has an activation function for defining its output. Also, the input of neuron is the sum of a bias and m points $(x_{1},\cdots ,x_{m})$ in which each of them are multiplied by $(w_{0},w_{1}, \cdots ,w_{m})$, respectively. If these weights and biases are chosen appropriately, a neuron can achieve to a better outcome. 

In general, a network has its specific number of layers and any layer has its specific number of neurons. Moreover, the number of inputs and outputs of network are based on the research topic, and in result it can be changed. In many situations, if the number of layers is two, an appropriate neural network can be designed. In this section, by using NGA-West2 databases provided by Berkeley \cite{databse}, we design a neural network. The parameters of studied data are as follows: \\

1-	Occurrence time in year

2-	The size of earthquake in Richter

3-	Epicenter's latitude in degree

4-	Epicenter's longitude in degree

5-	The focal depth in kilometer

6-	Epicentral distance in kilometer

7-	Hypocentral distance in kilometer\\

These seven parameters are shown by $a_{1},\cdots ,a_{7}$. We would design a neural network by one target and a set of six inputs. The target value is the severity of magnitude of earthquake $a_{2}$, and other parameters are network's input values. For obtaining better results, missing data are eliminated. The data matrix is ordered based on years from the oldest one. All the data are transferred to the $(-1,1)$ interval. Among all the data, ninety percent of it is chosen accidentally for training the neural network and remained ten per cent are used for testing it. 

Mean Square Error (MSE) is used for assessment of the error. For a vector $y=(y_{1},\cdots ,y_{n})$ and its approximation $\widehat{y}=(\widehat{y}_{1} ,  \cdots \widehat{y}_{n})$  , the MSE is as follows:

\begin{equation}
e=\dfrac{1}{n} \sum_{i=1}^{n} (y_{i} - \widehat{y}_{i})^{2}
\end{equation}

The used neural network has one input and one output layer by two hidden layers. The first hidden layer has 16 neurons and the second one has 24 neurons.  the program is written in a way that without changing the other parts, the number of neurons for hidden layers would be changeable. The activation function which is used for hidden layers are hyperbolic tangent sigmoid transfer functions (tansig) by following formulas: 

\begin{equation}
tansig(n)=\dfrac{2}{(1+e^{-2n})} -1
\end{equation}

In general, sigmoid is used for all the functions which have S shape. The activation function for the output layer which has only one neuron is pure linear transfer function.

\begin{figure}[!t] 
\centering
\includegraphics[width=3.3in]{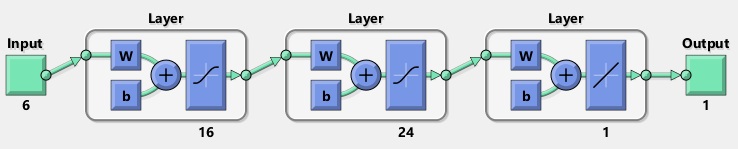}
\caption{The used neural network with two hidden layers}
\label{photo:1}
\end{figure}

The used neural network is shown in figure \ref{photo:1}. 

The last step is allocating weights and biases to the network. In order to do this, we should train the network in a way that the best weights and biases be chosen to minimize the MSE value. As this neural network has 6 input parameters and the number of neurons in the first hidden layer is 16, on the whole, we need 96 weights in this section. Although, as the number of neurons in the first hidden layer is 16 and in the next hidden layer is 26, we need 384 weights in this section to be aligned. Also, for moving from second layer to output, 24 weights are needed. The number of biases for the first layer is 16, 24 for the second layer, and one for the output. In this neural network, in sum, we need 545 weights and biases. We use ICA algorithm to find the best amount of them. In the next section the training method will be explained. 

As training any neural network leads to an optimization problem, one way to deal with it is using evolutionary optimization methods e.g. GA, Particle Swarm Optimization (PSO), and ICA. In this section, we will use ICA algorithm for determining neural network's weights and biases in training network. Also the result will be compared with GA-based network.  

To solve a problem by ICA algorithm, firstly an accurate definition of countries should be presented, and then cost function should be determined. In training the neural network, these two parameters would be defined as follows:\\

$\bullet$ 	$\textbf{Country:}$ the set of unknown parameters like biases and weights of neural network. A country plays the role of a neural network; in other words, it plays the role of a classifier.  \\

$\bullet$ 	$\textbf{Cost \ function:}$ refers to the mean square error of training data. 

The initial parameters of ICA algorithm are indicated in table \ref{table:1}. 
\renewcommand{\arraystretch}{1.7}
\begin{table}[h!]
\small
\caption{ICA algorithm initial parameters}
\begin{center}
\begin{tabular}{ |p{4.5cm}| | c | }

 \hline 
 \multicolumn{1}{|c||}{Parameter name}     &  \multicolumn{1}{|c||}{parameter value} \\
 \hline \hline
 The Number of  countries   & \multicolumn{1}{|c||}{1000}    \\
 \hline
 The number of primary empires  &   \multicolumn{1}{|c||}{100}  \\
 \hline
 The number decades (iterations) &  \multicolumn{1}{|c||}{200} \\
 \hline
\end{tabular}
\end{center}
\label{table:1}
\end{table}

\begin{figure}[!t]
\centering
\includegraphics[width=3.3in]{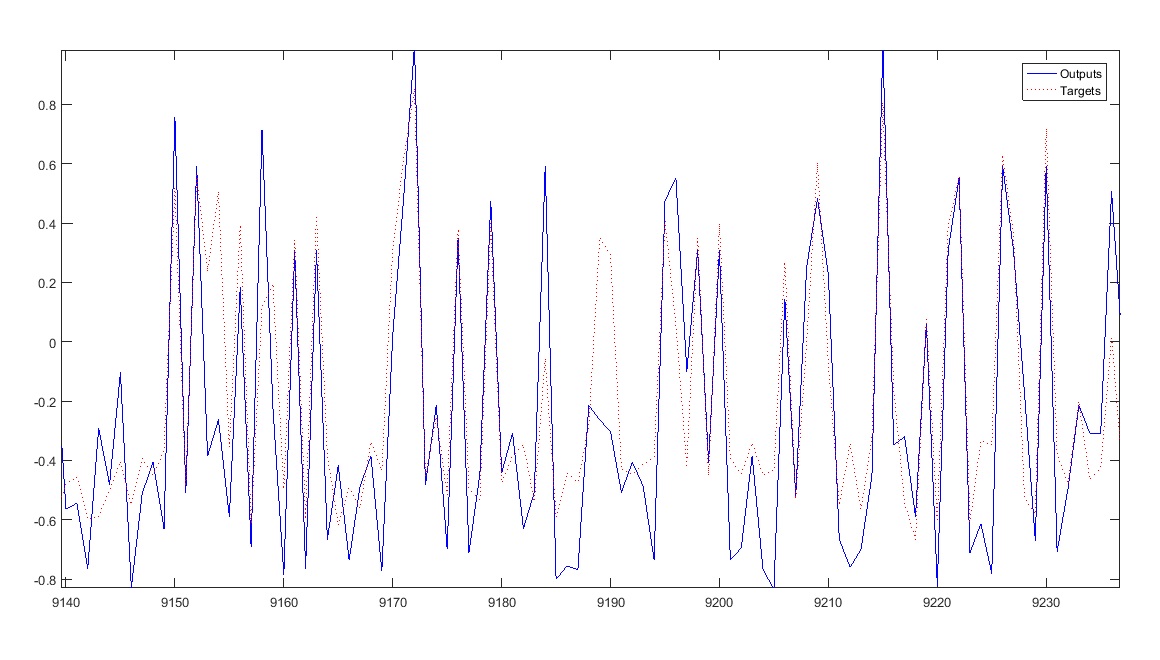}
\caption{Difference between the target data and the neural network output train data}
\label{photo:2}
\end{figure}

In used ICA algorithm, the number of iterations is 200 and the number of members of the initial population is 1000. The goal of algorithm is allocating best values for weights and biases in minimizing defined MSE for training data. The best achieved MSE for training data by using ICA algorithm is 0.097.\\
The difference between the target data and output of the training data in network when zoomed in on, has been indicated in the figure \ref{photo:2}.

\begin{figure}[!t]
\centering
\includegraphics[width=2.6in]{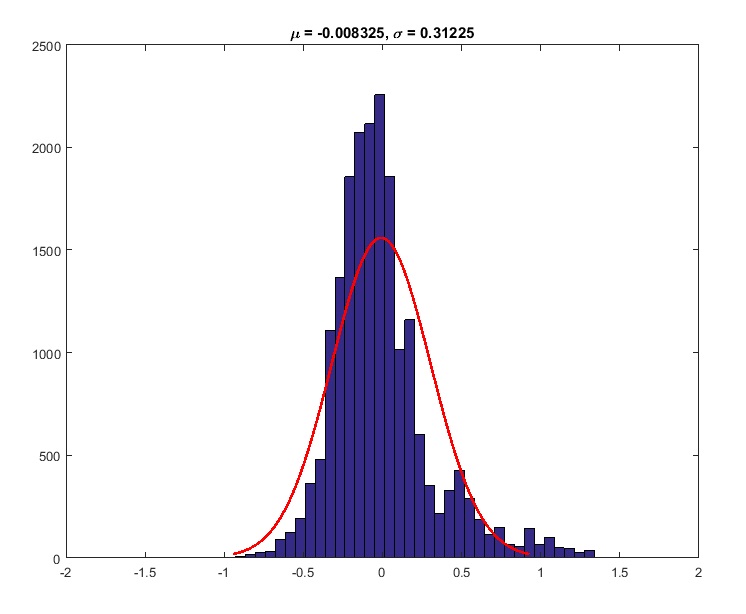}
\caption{Histogram of neural network training data error}
\label{photo:3}
\end{figure}

Average error for training data is equal to -0.008 and its variance is 0.312. The due histogram has been shown in the figure \ref{photo:3}. 

\begin{figure}[!t]
\centering
\includegraphics[width=3.5in]{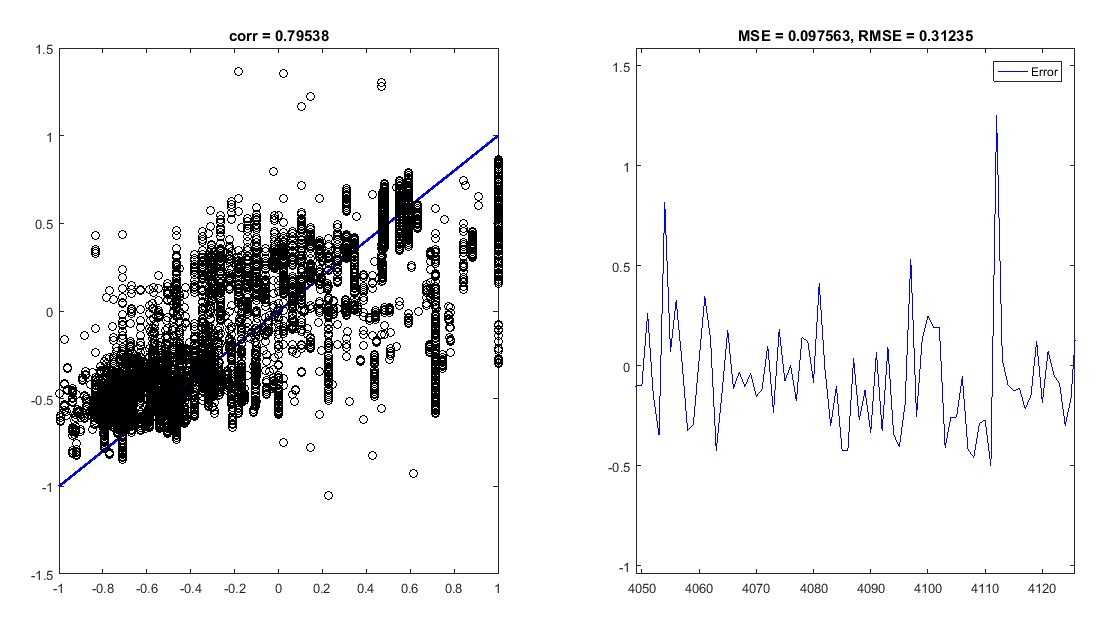}
\caption{Correlation between training data and the output of neural network with network error}
\label{photo:4}
\end{figure}

Correlation between training output data and the training target data is equal to 0.795, MSE=0.097, and RMSE=0.312 that RMSE is the square root of MSE. A portion of error changes and this correlation has shown in figure \ref{photo:4}. 

\begin{figure*}[!t]
\centering
\includegraphics[width=6.5in]{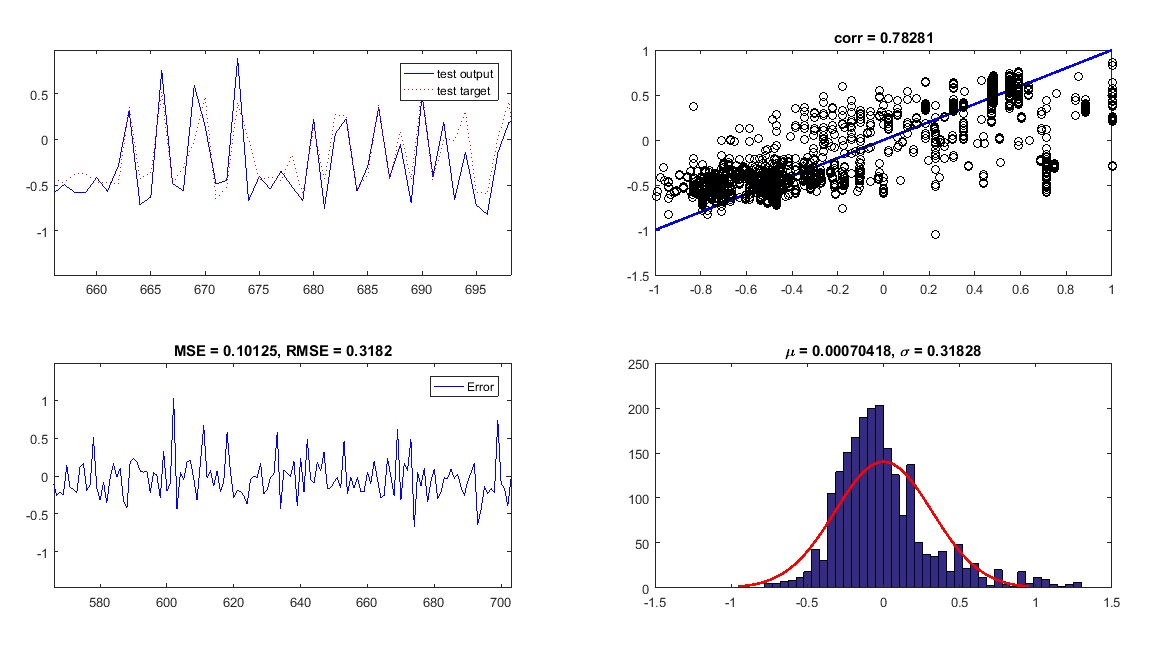}
\caption{test data evaluation of trained neural network by ICA algorithm}
\label{photo:5}
\end{figure*}

Similarly, the correlation, error function histogram, MSE, RMSE, and comparison of target values and the neural network output information of the test data has been shown in figure \ref{photo:5}. In this figure, for better vision of error figure and difference between test data and neural network output we zoomed in on. \\

Neural network testing reveals that for input data how much we can predict the magnitude and intensity of an earthquake by Richter. This prediction's MSE is equal to 0.101 and the correlation between this predicted values by using input parameters is equal to 0.782. 

\begin{figure*}[!t]
\centering
\includegraphics[width=6.5in]{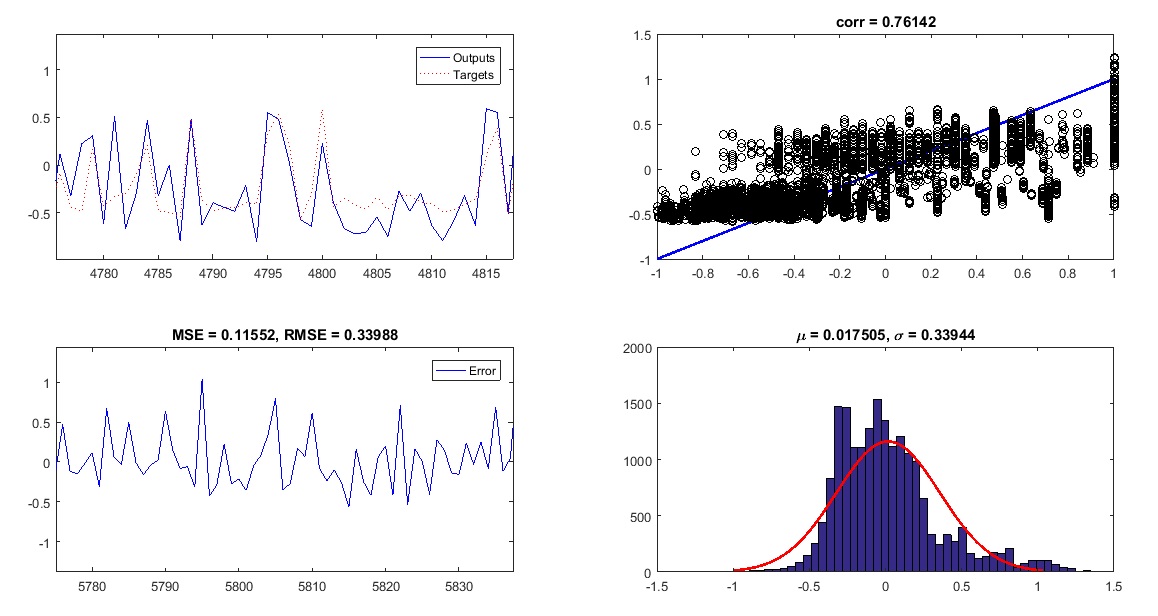}
\caption{test data evaluation of trained neural network by genetic algorithm }
\label{photo:6}
\end{figure*}

In a similar way, we trained the same neural network by genetic algorithm with equivalent structure as we did for ICA algorithm. For GA, the population size is 1000 and algorithm iteration is equal to 200. In any iteration 15 percent of data is selected as the elite population, 50 percent for crossover, and remained 35 percent for mutation. Similarly, the goal of algorithm is allocating the best weights and biases parameters to minimize MSE. After training the neural network by GA algorithm, it will be tested. The result for testing neural network is shown in figure \ref{photo:6}. 

As it can be seen in figure \ref{photo:6}, the obtained correlation is equal to 0.761 and MSE criteria for test data is 0.115. Therefore, training neural network by using ICA algorithms in figure \ref{photo:5} not only has a higher speed, but also it conducts better in training data by comparison with GA.

\section{Conclusion}

We have used MATLAB R2016b for designing a neural network for predicting intensity of earthquake. Neural network conducts very well in fitting functions and a neural network, also a simple one, can approximate very well any function. For defining the earthquake intensity prediction problem, we determined 6 input parameters as an input matrix, and then specified the earthquake intensity parameter as target variable. For assessing neural network, the MSE measure and regression analysis has been used. Designing artificial neural network in this way has not the ability for choosing parameters like the number of hidden layers. For this reason, after designing the primarily neural network with two hidden layers, the number of neurons of each layer and the kind of training functions has been changed to determine the effects of those parameters by using which ICA algorithms we allocate weights and biases values of the network and it has been compared with a neural network trained by GA. As we obtained, ICA was better than GA for training this network.

\section{Acknowledgment}
We take immense gratitude to Iran Seismological Center for providing a situation for making a research about application of neural network in seismology.

\end{document}